# WSSL: WEIGHTED SELF-SUPERVISED LEARNING FRAMEWORK FOR IMAGE-INPAINTING

Shubham Gupta, Rahul Kunigal Ravishankar, Madhoolika Gangaraju, Poojasree Dwarkanath
and Natarajan Subramanyam
*PES University*
*100 Feet Ring Road, Banashankari Stage III, Bengaluru, Karnataka, India*

**ABSTRACT**

Image inpainting is the process of regenerating lost parts of the image. Supervised algorithm-based methods have shown excellent results but have two significant drawbacks. They do not perform well when tested with unseen data. They fail to capture the global context of the image, resulting in a visually unappealing result. We propose a novel self-supervised learning framework for image-inpainting: Weighted Self-Supervised Learning (WSSL) to tackle these problems. We designed WSSL to learn features from multiple weighted pretext tasks. These features are then utilized for the downstream task, image-inpainting. To improve the performance of our framework and produce more visually appealing images, we also present a novel loss function for image inpainting. The loss function takes advantage of both reconstruction loss and perceptual loss functions to regenerate the image. Our experimentation shows WSSL outperforms previous methods, and our loss function helps produce better results.

## 1. INTRODUCTION

Image Inpainting involves reconstructing scratches, splotches, blurred or discolored regions of an image using precise algorithms. Several Image inpainting methods based on sequential algorithms (Anitha & S Natarajan 2008), Convolutional Neural Networks (CNNs) (Wang 2018), and Generative Adversarial Networks (GANs) (Demir, U. & Unal, G. 2018) have been utilized in the past to perform image inpainting. However, these methods produce unsatisfactory results when the regions are too big, as seen in (Anitha and S Natarajan 2008), (Wang 2018), or require access to expensive resources (Demir, U. & Unal, G. 2018). A common factor among these methods is that they are fully supervised algorithms.

Self-supervised learning (SSL) algorithms have recently surpassed supervised algorithms in Computer Vision tasks like object classification and detection (Chen et al 2020), (Grill et al 2020). Taking inspiration from their success in learning feature representations without manually annotated data, we employ self-supervised learning for image-inpainting. A self-supervised algorithm has a pretext training phase and a downstream training phase. During the pretext training, the model learns the general feature representations such as texture, boundaries, and shape of the subjects present in the image. It does so by learning pseudo-labels for very simple tasks like rotation prediction (Gidaris, S., Singh, P. & Komodakis, N. 2018), and image-colorization (Zhang, R., Isola, P. & Efros, A.A. 2017), etc.

We study the effectiveness of single and combined pretext tasks in our work. We also vary the weights of pretext tasks in combinations to find a trend. To study the effectiveness of different weighted combinations, we build an SSL framework called Weighted Self-supervised Learning (WSSL). We can train on multiple pretext tasks and assign weights to each of the individual tasks through our framework. Assigning weights to the pretext tasks helps vary the importance given to different features learned by the individual pretext tasks. While the features learned during training are essential, the loss function we optimize while training is often overlooked (H. Zhao, O. Gallo, I. Frosio & J. Kautz, 2017). For this reason, we study multiple reconstruction loss functions. These loss functions may achieve an acceptable score quantitatively, but it fails to reconstruct visually appealing images (H. Zhao, O. Gallo, I. Frosio & J. Kautz, 2017). We also explore perceptual loss functions in combination with reconstruction loss to create a balanced novel loss function. Our results show that the presented loss function significantly improves the performance of our SSL framework. To summarize our contributions in this paper:

1. We present a novel self-supervised learning framework for image inpainting, Weighted Self-supervised Learning (WSSL).
2. We compare the performance of various pretext tasks under different weighted configurations.
3. We present a novel loss function that utilizes both reconstruction and perceptual loss. The resulting function, when optimized, significantly boosts the performance of our framework.

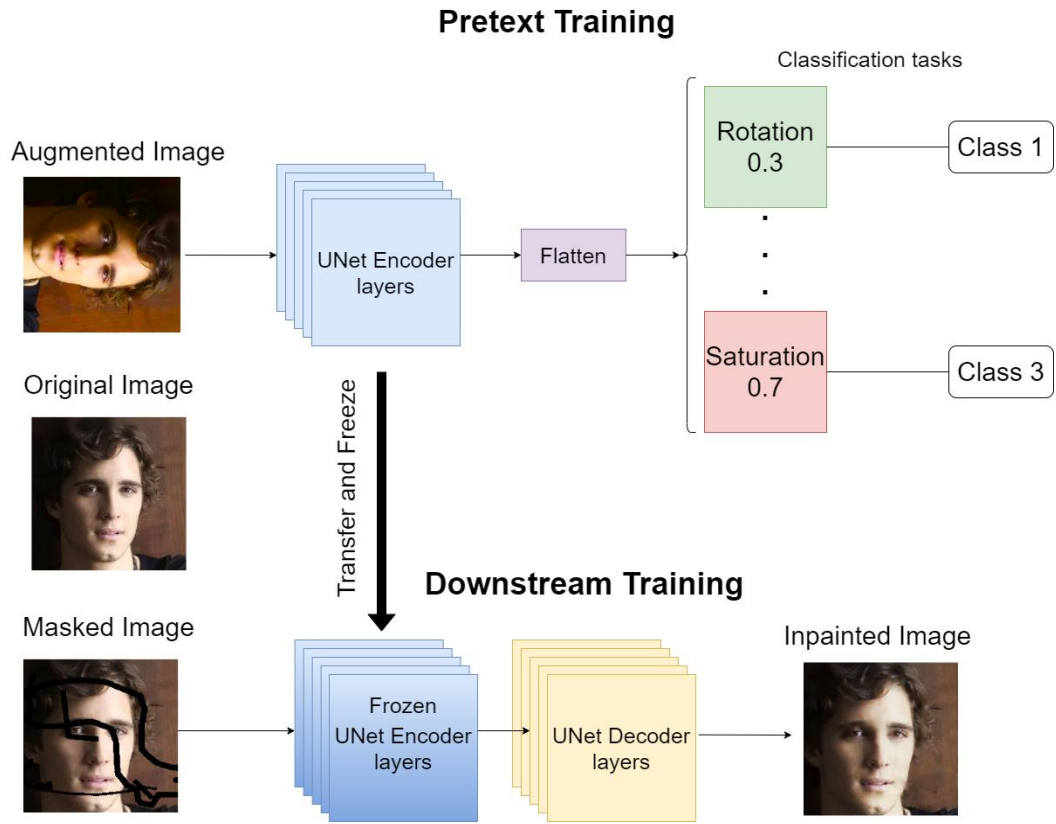

Figure 1. The proposed architecture of the Weighted Self-Supervised Learning (WSSL) framework demonstrates a weighted double pretext task. The Rotation classification task has a loss weight of 0.3 and the Saturation classification task has a loss weight of 0.7

## 2. LITERATURE SURVEY

(Elharrouss et al 2020) elaborates on several Sequential algorithms, methods that use CNNs, and methods that use GANs. The two categories of sequential algorithms include patch-based algorithms and diffusion-based algorithms. In patch-based methods, the missing regions are filled patch-wise with the best match (Ruzic, Tijana & Aleksandra Pižurica, 2015), (K. H. Jin & J. C. Ye 2015) from the known regions of the image. In diffusion-based methods, the image content is smoothly propagated from the boundary of the missing region to its interior (Li, Haodong, Weiqi Luo & Jiwu Huang 2017) (Kangshun et al 2014).

The global structure of the image can be captured using CNNs or encoder-decoder networks. (Pathak et al 2016) proposed a network called Context Encoders that has an encoder-decoder based architecture with the encoder being able to learn the semantically meaningful representation of the images and the decoder being able to synthesize high-level features for image inpainting. GANs have a contextual attention module that is helpful for inpainting. Although the GAN-based methods perform better than CNNs in many cases (Shin, Y.G.

et al 2020), (Vitoria, P.; Sintes, J. & Ballester, C 2019), it is computationally expensive, and the training time is high.

(H. Zhao, O. Gallo, I. Frosio & J. Kautz 2017) discuss the variety of loss functions that are overlooked when solving the image restoration problem. Their work shows the shortcomings of the L2 loss function, which only penalizes enormous errors and assumes the pixels follow a gaussian distribution. This behaviour is highly ideal and very unlikely. Instead, they look into perceptual loss functions such as the Structural Similarity Index (SSIM) and Multiscale Structural Similarity Index (MS-SSIM). Their results show that using L1 and MS-SSIM loss results in a consistent image reconstruction with minimal colour blemishes.

(Gidaris, S., Singh, P. & Komodakis, N. 2018) proposed a simple yet effective way of learning image representation by predicting image rotations. (S. Yenamandra et al 2020) proposed a DNN framework that can learn both in self-supervised and semi-supervised modes that leverage the information available within the incomplete images. Removing some of the known regions and aiding the DNN framework to learn to inpaint the removed region enables the DNN framework to learn to inpaint even without corresponding uncorrupted images.

Inpainting involves the extraction and generalization of the structural characteristics of images, such as texture and border. (Shin, Y.G. et al 2020), (Yamaguchi, S.Y. et al 2021) propose a network that captures information regarding the shape and texture of objects in images by predicting the rotation, sharpness, and saturation of the images simultaneously. Rotation, sharpness, and saturation capture the object's general shape, edges, and colour patterns respectively.

Furthermore, self-supervised learning on multiple pretext tasks has shown to be robust to out-of-domain samples (Albuquerque, I. et al 2020). Their work proposes a two-stage network for predicting the output for the Gabor filter (Fogel, I. & Sagi, D. 1989) on an image using either a single or multiple tasks. The result obtained easily competes against supervised methods on domain generalization and outperforms in localization. Through this, we are inspired to build WSSL's two-stage architecture with variations in pretext tasks.

## 3. METHODOLOGY

### 3.1 Data

We use the CelebA HQ dataset (Karras, T. et al 2017) which contains 30,000 high-resolution images of celebrities, each of 1024x1024 pixel resolution. We resize the images to 256 x 256 pixels and then obtain a random crop of size 224 x 224 pixels. We use a predefined dataset of masks, Quick Draw Irregular Mask (QD-IMD) (Iskakov, K. et al 2018) dataset. We filtered the masks such that the percentage of missing pixels is within 30 - 50%. The masks and images from the dataset are paired randomly during the training phase.

### 3.2 Weighted Pretext Tasks

The augmentations considered in this paper are rotation, saturation, and sharpness. The combination of pretext tasks used for testing our framework includes rotation, saturation, sharpness, rotation-saturation, rotation-sharpness, saturation-sharpness, and rotation-saturation-sharpness. In the case of multiple pretext tasks, intuitively, the features learned by one of the tasks might give a better semantic representation of the image than the other. Assigning weightage to each of these tasks based on their performance can help distribute the importance of the features that these tasks have identified. The weightage would, in turn, lead to much better feature representations. Hence, we experiment with combinations of pretext tasks, with different weights assigned to each pretext task. The image is discretized into four rotations of 0, 90, 180, and 270 degrees. The sharpness and saturation of the image are discretized into 0, 0.25, 0.75, and 1.0.

## 3.3 Loss Function for Downstream Task

### 3.3.1 SSIM

SSIM stands for Structural Similarity Index Measure and is used as a metric to measure the similarity between two given images using a value between 0 and 1. SSIM incorporates critical perceptual phenomena such as luminance and contrast masking and is a perception-based model (Zhou et al 2004).

$$SSIM = \frac{(2\mu_x\mu_y + c_1).(2\alpha_{xy} + c_2)}{(\mu_x^2 + \mu_y^2 + c_1).(\alpha_x + \alpha_y + c_2)}$$

$$L_{SSIM} = 1.0 - SSIM$$

### 3.3.2 Log(cosh)

Log(cosh) is the logarithm of the hyperbolic cosine of the prediction error. It is a reconstruction loss function. Log(cosh(x)) is approximately equivalent to $\frac{x^2}{2}$ for small values of x and $|x| - \log(2)$ for larger values of x. log(cosh), therefore, works mostly like the mean squared error, but will not be so strongly affected by the occasional wildly incorrect prediction (Moshagen, T., Adde, N.A. & Rajgopal, A.N. 2021). This is the advantage of log(cosh) over mean squared error. The log(cosh) function behaves as the L1 loss for small values and like L2 loss for larger values, giving us the best of both worlds.

$$L_{log(cosh(a,t))} = \frac{1}{a} log(\frac{e^{at} + e^{-at}}{2})$$

### 3.3.3 Novelty

Reconstruction losses perform well quantitatively but fail to regenerate a visually pleasing image. A metric like SSIM, however, will help produce a less aberrated image. For this reason, we implement a weighted sum of SSIM and log(cosh). The novel loss function is a weighted sum of SSIM and log(cosh). The weight $\alpha$, is a free parameter that varies based on the application. $\alpha = 0.84$ gave us optimal results for Image Inpainting.

$$L_{total} = \alpha.L_{log(cosh)} + (1 - \alpha).L_{SSIM}$$

where, $\alpha = 0.84$

## 4. IMPLEMENTATION DETAILS

Our entire codebase is written in Python 3.9 using Tensorflow 2.0 framework. We utilize a vanilla UNet architecture (Ronneberger, O., Fischer, P. & Brox, T. 2015) which consists of five encoders and five decoders. The architecture is designed to regenerate images of size 224 x 224 pixels. The pretext model consists of UNet encoders, followed by different branches of classification layers depending on the number of pretext tasks used. For example, if a combination of two pretext tasks is being used, there would be two different branches, each with its classification layers. The pretext model was trained for 40 epochs using an Adam optimizer with a learning rate of $1e - 4$, a batch size of 64. For the downstream model, the first 16 layers of the pretext model are frozen to keep the learned feature representations, and the classification layers are replaced with the decoder layers of the UNet model to perform image-inpainting. The downstream model uses the same settings for 20 epochs. The models were trained using Nvidia's K80 GPU. Our best model, further discussed in Section 5, took 4 hours to train including pretext and downstream tasks.

## 5. RESULTS AND DISCUSSION

In this section, we compare our image inpainting method with other methods that have been proposed recently such as ShiftNet (Yan, Z. et al 2018) and PIC (Zheng, C., Cham, T.J. and Cai, J., 2019). We evaluate all methods quantitatively and qualitatively using Structural Similarity Index Measure (SSIM) and Peak Signal to Noise Ratio (PSNR) metrics.

Table 1. Results using WSSL with a single pretext task

| Pretext task | SSIM | PSNR |
|---|---|---|
| Sharpness | 0.9281 | 25.84 |
| Saturation | 0.9596 | 29.90 |
| **Rotation** | **0.9608** | **29.89** |

We have used simple pretext tasks such as Rotation, Saturation, and Sharpness to perform our experiments. Rotation forces the model to learn the general shape of the objects within the image. Altering the sharpness of the image changes the emphasis put on the edges of the images' objects. Altering the saturation of the images will force the model to look for patterns and correlate colour with the objects present within the image. In Table 1, we can see that the network can learn to extract a good set of features with the pretext task Rotation. We can see that just identifying the edges or colours and patterns within the image is not sufficient to inpaint the image.

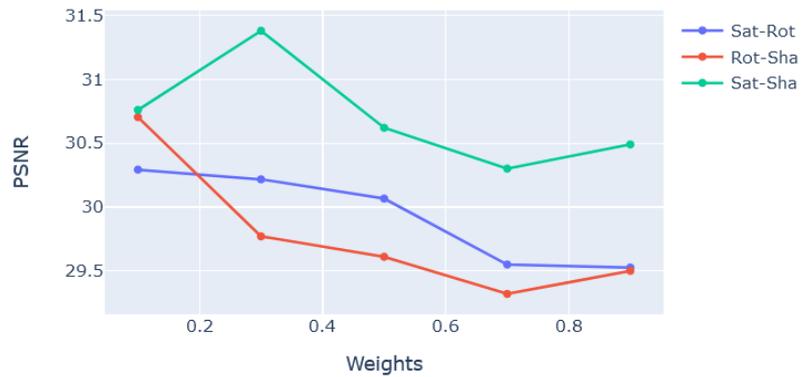

(a) PSNR vs Weights of the first pretext task.

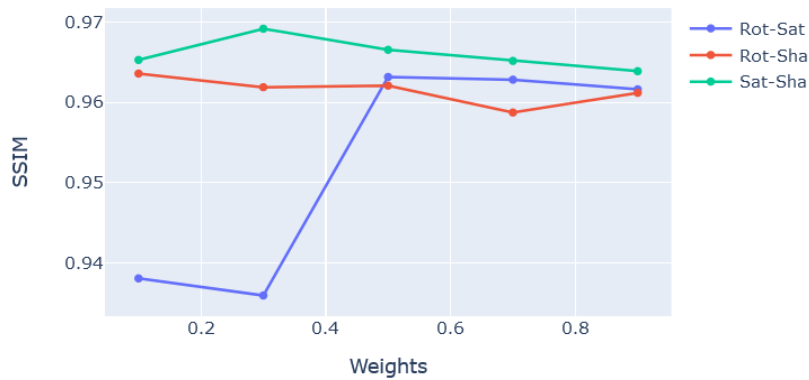

(b) SSIM vs Weights of the first pretext task.

Figure 2. Finding trends in performance of WSSL with double pretext task

We have used combinations of two or more pretext tasks and weighted each pretext task, thereby varying the importance of identifying shapes, edges, and colour patterns. In Table 2, when using a combination of Rotation and Saturation, we can see that the network performs better with a higher weightage given to the Rotation pretext task and a lower weightage given to saturation, as shown in Figure 2.

Table 2. Results using WSSL with double pretext task

| Pretext task | Loss weights | | SSIM | PSNR |
|---|---|---|---|---|
| Rotation - Sharpness | 0.9 | 0.1 | 0.9612 | 29.50 |
| | 0.7 | 0.3 | 0.9587 | 29.31 |
| | 0.5 | 0.5 | 0.9621 | 29.61 |
| | 0.3 | 0.7 | 0.9619 | 29.77 |
| | **0.1** | **0.9** | **0.9636** | **30.70** |
| Rotation - Saturation | 0.9 | 0.1 | 0.9616 | 30.29 |
| | 0.7 | 0.3 | 0.9628 | 30.21 |
| | **0.5** | **0.5** | **0.9631** | **30.06** |
| | 0.3 | 0.7 | 0.9359 | 29.55 |
| | 0.1 | 0.9 | 0.9380 | 29.52 |
| Saturation - Sharpness | 0.9 | 0.1 | 0.9639 | 30.49 |
| | 0.7 | 0.3 | 0.9651 | 30.30 |
| | 0.5 | 0.5 | 0.9665 | 30.62 |
| | **0.3** | **0.7** | **0.9692** | **31.38** |
| | 0.1 | 0.9 | 0.9653 | 30.76 |

Table 3. Results using WSSL with triple pretext task

| Pretext task | Loss weights | | | SSIM | PSNR |
|---|---|---|---|---|---|
| Saturation - Sharpness - Rotation | 0.6 | 0.2 | 0.2 | 0.9667 | 31.03 |
| | 0.2 | 0.2 | 0.6 | 0.9634 | 30.23 |
| | 0.2 | 0.6 | 0.2 | 0.9662 | 30.74 |
| | **0.3** | **0.4** | **0.3** | **0.9685** | **31.04** |

On the other hand, we can see that the network performs better with a lower weightage given to the Rotation pretext task when a combination of Rotation and Sharpness pretext tasks are used. Nevertheless, the network performs better when a higher weightage is given to the sharpness pretext task when using a combination of saturation and sharpness. Therefore, the weight given for each pretext task is a free parameter that depends on the combination of pretext tasks used.

In Figure 4, we can see that the mean of the SSIM obtained with a combination of two pretext tasks is much higher than the individual pretext tasks. By combining all three pretext tasks, we can see that the SSIM further increases. We can also see that the standard deviation of the SSIM decreases, and the results are more consistent with an increase in the number of pretext tasks (Table 3). Therefore, combining the pretext tasks would span a broader feature space, leading to a better feature representation of the images.

As shown in Table 4, WSSL with the custom loss function performs better when compared to using only SSIM or only log(cosh) as loss functions. It can also be observed from the PSNR and SSIM values, and Figure 3, that our model outperforms Shiftnet and PIC models. PIC generates a semantically meaningful result whereas our framework generates an optimal result using a single ground-truth image. Hence, a higher SSIM value is obtained through our framework although the results from PIC look more realistic.

Table 4. Performance of models on CelebA dataset.

| Method | SSIM | PSNR |
|---|---|---|
| ShiftNet | 0.7878 | 20.88 |
| WSSL + log(cosh) loss function (ours) | 0.9338 | 29.46 |
| PIC | 0.9434 | 31.27 |
| WSSL + SSIM loss function (ours) | 0.9629 | 28.90 |
| **WSSL + Novel loss function (ours)** | **0.9692** | **31.38** |

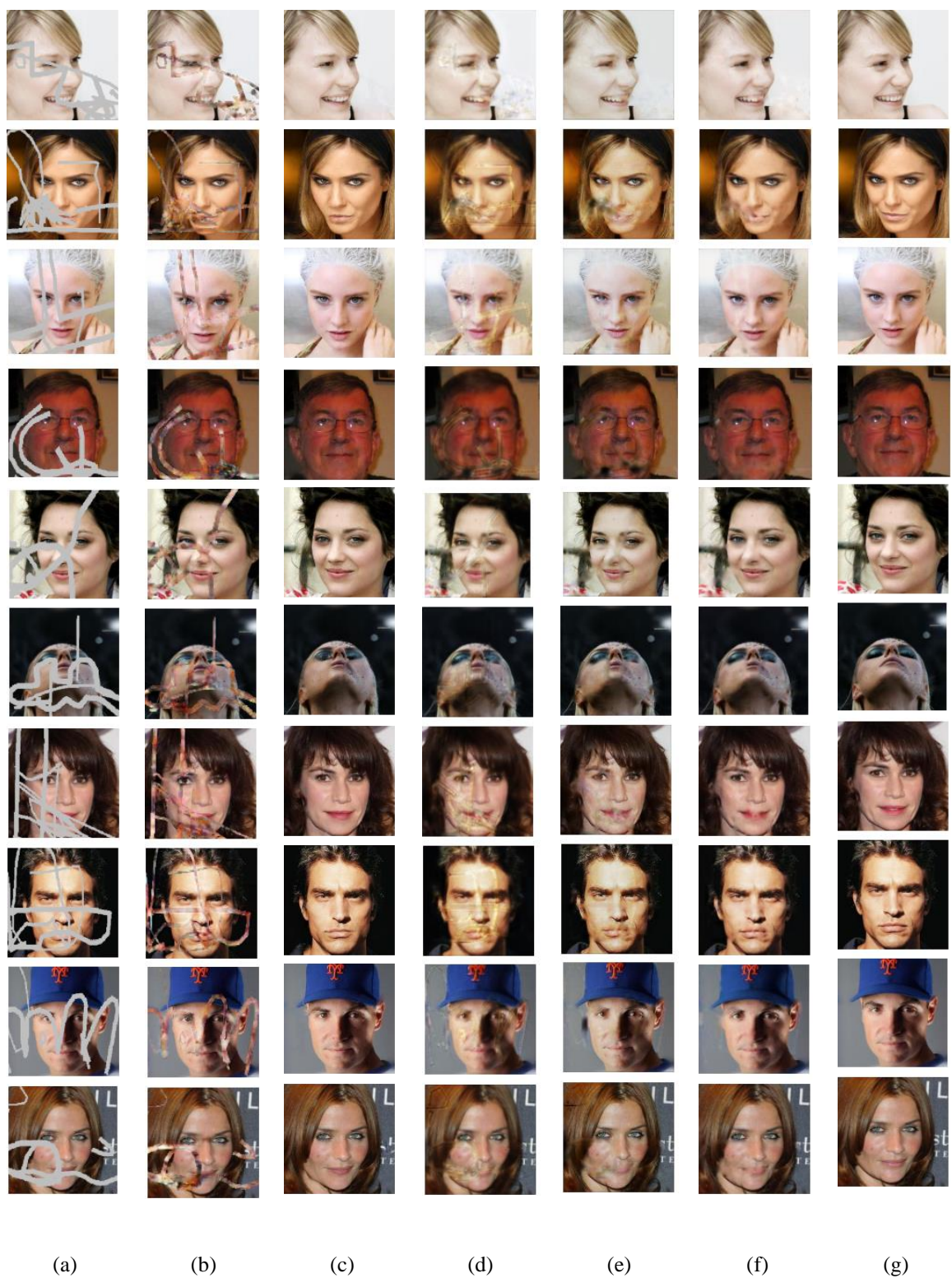

(a)	(b)	(c)	(d)	(e)	(f)	(g)

Figure 3. (a) Masked image (b) Shiftnet (Yan, Z. et al 2018) (c) PIC (Zheng, C., Cham, T.J. and Cai, J., 2019) (d) WSSL + log(cosh) loss function (e) WSSL + SSIM loss function (f) WSSL + novel loss function (g) Original image

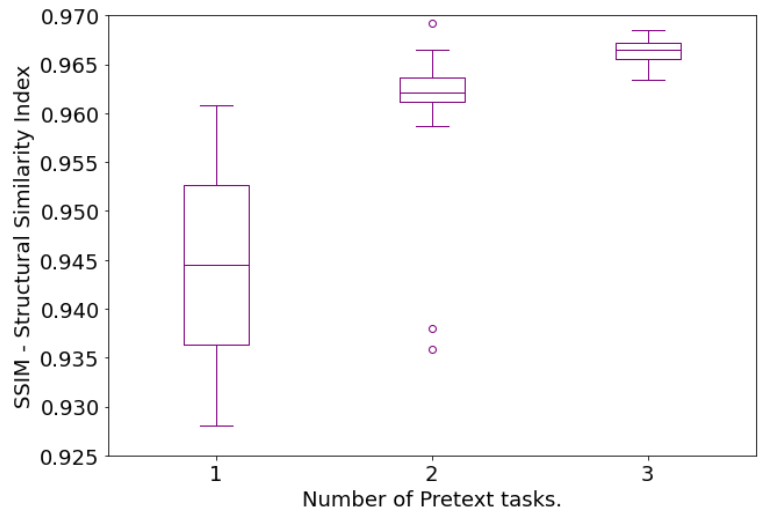

Figure 4. Plot comparing SSIM obtained from networks trained with single, double, and triple pretext tasks

## 6. CONCLUSION

In this paper, we have demonstrated the advantages of Self-supervised learning, such as improving robustness, better image feature representation, domain generalization, and transfer learning. Our experiments show that simple augmentations provide good feature representations of the images. We demonstrated the effectiveness of our novel framework, Weighted Self-supervised Learning (WSSL), through different weighted combinations of pretext tasks for Image Inpainting. Providing weights to the pretext tasks helps guide the framework to learn better feature representations of an image. Using multiple pretext tasks reduces the variation in the downstream results. The analysis of the compatibility of different pretext tasks and how they improve performance when utilized together can be a part of future work.

We also study the performance of different loss functions for image-inpainting. Our observation is that a weighted combination of reconstruction and perceptual loss functions is best suited to produce a visually appealing result. WSSL combined with our novel loss function significantly boosts the inpainting performance. We observe that our loss function's inpainted results have more recovered structures and colour than the other loss functions. We could further test the efficacy of WSSL on other image generation tasks in the future, such as image segmentation, object detection, and image super-resolution.